\documentclass[manuscript]{acmart}

\usepackage{todonotes}
\usepackage{amsmath}
\usepackage{subcaption}

\AtBeginDocument{%
  \providecommand\BibTeX{{%
    \normalfont B\kern-0.5em{\scshape i\kern-0.25em b}\kern-0.8em\TeX}}}

\setcopyright{acmcopyright}
\copyrightyear{2018}
\acmYear{2018}
\acmDOI{10.1145/1122445.1122456}

\acmConference[FDG'20]{FDG'20: Foundations of Digital Games}{September 15--18, 2020}{Malta}
\acmBooktitle{FDG'20: Foundations of Digital Games,
 September 15--18, 2020, Malta}
\acmPrice{15.00}
\acmISBN{978-1-4503-XXXX-X/18/06}

\copyrightyear{2020}
\acmYear{2020}
\setcopyright{acmlicensed}\acmConference[FDG '20]{International
Conference on the Foundations of Digital Games}{September 15--18,
2020}{Bugibba, Malta}
\acmBooktitle{International Conference on the Foundations of Digital
Games (FDG '20), September 15--18, 2020, Bugibba, Malta}
\acmPrice{15.00}
\acmDOI{10.1145/3402942.3402954}
\acmISBN{978-1-4503-8807-8/20/09}

\begin{document}

\title{Mech-Elites: Illuminating the Mechanic Space of GVG-AI}

\author{M Charity}
\affiliation{%
  \institution{New York University}
  \streetaddress{370 Jay Street}
  \city{Brooklyn}
  \state{NY}
  \postcode{11201}
}
\email{mlc761@nyu.edu}

\author{Michael Cerny Green}
\affiliation{%
  \institution{New York University; OriGen.AI}
  \streetaddress{370 Jay Street}
  \city{Brooklyn}
  \state{NY}
  \postcode{11201}
}
\email{mike.green@nyu.edu}

\author{Ahmed Khalifa}
\affiliation{%
  \institution{New York University}
  \streetaddress{370 Jay Street}
  \city{Brooklyn}
  \state{NY}
  \postcode{11201}
}
\email{ahmed@akhalifa.com}

\author{Julian Togelius}
\affiliation{%
  \institution{New York University}
  \streetaddress{370 Jay Street}
  \city{Brooklyn}
  \state{NY}
  \postcode{11201}
}
\email{julian@togelius.com}

\renewcommand{\shortauthors}{Charity et al.}

\begin{abstract}
This paper introduces a fully automatic method of mechanic illumination for general video game level generation. Using the Constrained MAP-Elites algorithm and the GVG-AI framework, this system generates the simplest tile based levels that contain specific sets of game mechanics and also satisfy playability constraints. We apply this method to illuminate the mechanic space for four different games in GVG-AI: Zelda, Solarfox, Plants, and RealPortals. With this system, we can generate playable levels that contain different combinations of most of the possible mechanics. These levels can later be used to populate game tutorials that teach players how to use the mechanics of the game.
\end{abstract}



\maketitle

\section{Introduction}

Video game levels are one of the most important assets of gaming. They provide spaces where players experience the game, the environments where they interact with the entire range of the possible mechanics. From a player's perspective, a game mechanic is ``...everything that affords agency in the game world''~\cite{sicart2008defining}. For example, a jumping action in a platforming game would be considered a game mechanic, as would recieving an item in a role-playing game or destroying a monster in an action game. Levels often showcase different game mechanics, either in an isolated setting to help the player hone a particular mechanic - a type of tutorial called a carefuly designed experience~\cite{green2017press} - or in combination with many other mechanics to glean the complex interplay between them. 
 
While simpler games only require a few tutorial environments or levels to teach a player their mechanics, more complex games might need more specific and niche levels in order to effectively teach the player without overwhelming them. Generating these tutorial levels by hand and identifying the critical mechanics needed to play the game can be tedious for the developer and ultimately may not be beneficial to the player if a more suitable tutorial level can be created to achieve the same goal. A good baseline tutorial level is one that is simple and direct with its mechanic demonstration. The AtDelfi project~\cite{green2018atdelfi} mentioned automated ``experience generation'' as a future goal of tutorial generation, and this project was primarily motivated to help fill this need.

In this paper, we introduce the use of AI methods to identify the mechanics of a game and generate levels from this list of mechanics. We are searching the space to find levels that both demonstrates individual critical mechanics and the combination of them. The evolved levels are playable with respect to gameplaying agents and they contain simple layouts and straightforward designs. This has been done in the past as a proof of concept for Mario levels, \cite{khalifa2019intentional}; where mechanics were predefined for the algorithm. We apply that method with some modifications, to 4 different games from the GVG-AI framework~\cite{perez2019general}: Zelda, Solarfox, Plants, and RealPortals. In this work, the game mechanics are automatically parsed from a game's description file instead of being fed by the designer. One potential application of this level generation would be to augment a tutorial generation system, such as AtDelfi~\cite{green2018atdelfi}, so that it can automatically develop levels that teach the different game mechanics. 


\section{Background}

Search-based PCG is a technique that uses search methods to find game content~\cite{togelius2011search}. Evolutionary algorithms are a class of stochastic optimization methods popularly used for PCG. Such algorithms programmatically apply concepts from Darwinistic evolutionary theory, such as mutation, population, and trans-generational genetic heritage, to find optimized solutions.
The Feasible Infeasible 2-Population (FI2Pop) genetic algorithm is one such algorithm that uses a dual-population technique~\cite{kimbrough2008feasible}. The ``feasible'' population attempts to improve the overall quality of solutions, or ``chromosomes,'' contained within. The other population - the ``infeasible'' population - attempts to satisfy a group of constraints that will move the ``chromosomes'' containted within to the ``feasible'' population. During evolution, ``chromosomes'' move between both populations whenever they satisfy or break these constraints.

Quality diversity (QD) algorithms are a class of methods that fall under the evolutionary umbrella. QD allows for a simultaneous focus on the quality of results in addition to maintaining diversity using explicit; separating it from traditional multi-objective optimization strategies, and making it a great candidate for PCG~\cite{gravina2019procedural}. The MAP-Elites (ME) algorithm~\cite{mouret2015illuminating} is one such QD algorithm that maintains a map of $n$-dimensions in place of a population. The elites are sampled to recreate a competing younger generation, which try to replace the older generation. These dimensions correspond to unique behavioral characteristics or traits that can help differentiate between different individuals. Example characteristics can include the number of enemies in a level, the solution length of a puzzle, etc. PlayMapper~\cite{warriar2019playmapper} used MAP-Elites algorithm for the Mario AI Framework~\cite{togelius20102009}. The system illuminates the level space based on player specific features and level specific features. Overall, the system showed how search-based PCG technique can be more effectively used by game designers.

Constrained MAP-Elites (CME)~\cite{khalifa2018talakat} is a hybrid genetic algorithm that combines the FI2Pop constrained optimization algorithm with MAP-Elites. Within each cell are stored two populations (``feasible'' and ``Infeasible''). Chromosomes can be moved between cells (if their dimensions shift) and/or between populations within their cell (if they successfuly outgrow their constraints or fail to do so). Constrained MAP-Elites allows a complex quality diversity search to optimize toward a given problem, making it a useful tool for PCG.

Khalifa et al.~\cite{khalifa2018talakat} used Constrained MAP-Elites to develop a range of level types for bullet-hell games, by characterizing levels based on the strategy required and dexterity a player would need to be successful. The Evolutionary Dungeon Designer project~\cite{alvarez2019empowering} uses Constrained MAP-Elites to allow mixed-initiative dungeon design. Users can tune the dimension settings to their liking in order to generate interesting level layouts. Most similar to this work, Constrained MAP-Elites has been used to generate mini-levels, or ``scenes,''~\cite{khalifa2019intentional} in the Mario AI Framework, by mapping mechanics triggered during gameplay.

Several research projects have attempted to generate game levels targeted to explore different dimensions of level space. Ashlock et al.~\cite{ashlock2010automatic,ashlock2011search} explored different evolutionary techniques for puzzle generation of various difficulties.  Jennings et al. built a system which dynamically constructs levels to be appropriately challenging to the player~\cite{jennings2010polymorph}. \textit{Refraction} (Center for Game Science at the University of Washington 2010) generates levels that showcase certain features (which in turn are associated with certain mechanics)~\cite{smith2012case}. 

\section{General Video Game Artificial Intelligence (GVG-AI)}\label{sec:GVG-AI}
The GVG-AI framework is a platform for automatic general video game research~\cite{perez2019general}. The framework provides multiple tracks including game playing~\cite{perez2016general}, level generation~\cite{khalifa2016general}, learning~\cite{torrado2018deep}, rule generation~\cite{khalifa2017general}, and two-player gameplay~\cite{gaina2016general}. 

Every game in the GVG-AI Framework is described in Video Game Description Language (VGDL)~\cite{ebner2013towards}. VGDL is encompassing enough to describe a wide variety of simple 2D games, yet remains easy to read for humans. Some of them are adaptations of known games, such as \emph{pacman} (Namco, 1980), \emph{Plants vs Zombies} (PopCap Games 2009), and \emph{Galaga} (Namco 1981), others are demakes of big games, such as \emph{The Legend of Zelda} (Nintendo, 1986) and \emph{Pokemon} (Game Freak, 1996), while others are brand new games, such as \emph{Wait for Breakfast} where the player must remain idle in order to win; a game that is difficult for artificial players to solve.

A VGDL game consists of two file types: the game description file and one or more level files. Four parts make up the game description file: a \emph{Sprite Set} which determines which game objects exist and how they look and behave, an \emph{Interaction Set} which describes how sprites interact, a \emph{Termination Set} to dictate how the game ends, and finally a \emph{level mapping} between game sprites and their ASCII representation in the level files.

The four GVG-AI games used in this work are Zelda, Solarfox, Plants, and RealPortals. Each was selected based on a previous work~\cite{bontrager2016matching}, which categorized GVG-AI games based on how they were played. These four games contain a diverse array of mechanics, terminal conditions (time-based (Plants), lock-and-key (Zelda and RealPortals), and collection (SolarFox)), and aggregately incorporate ranging levels of complexity. For example, whereas Zelda is a relatively simple lock-and-key game, RealPortals requires complex problem-solving, and Plants contains relatively enormous maps to search. Thus, we selected these as a representative set of the GVG-AI framework's games.
\begin{itemize}
    \item \textbf{Zelda:} is a GVG-AI adaptation of the dungeon system in \emph{The Legend of Zelda} (Nintendo 1986). The player must pick up a key and unlock a door in order to beat a level. Monsters populate the level and can kill the player, causing them to lose. The player can swing a sword, which can destroy monsters and grant points.
    \item \textbf{Solarfox:} is a GVG-AI adaptation of \emph{Solar Fox} (Bally/Midway Mfg. Co 1981). The player must dodge both enemies and their flaming projectiles in order to collect all the ``blibs'' in the level. The player gains a point for each blib collected, and victory is granted after collecting all blibs in the level. Several levels contain ``powered blibs,'' which are worth no points. If a player collides with a powered blib, it will spawn a ``blib generator,'' which as the name implies, can spawn more blibs to collect and gain more points. If a player touches a blib generator, however, the generator will be destroyed and no longer generate any more blibs. Good gameplay invokes a balance of short- and long-term strategy, balancing the greed of winning the level against getting more points and risking loss.
    \item \textbf{Plants:} is a GVG-AI adaptation of \emph{Plants vs. Zombies} (PopCap Games 2009), a tower defense-style game. If the player survives for $1000$ game ticks, they win. Zombies spawn on the right side of the screen and move left, and the player loses if a zombie reaches the left side. Plants, which the player must grow in specific ``marsh'' tiles, can destroy zombies by automatically firing zombie-killing peas. Each zombie killed is worth a point. Occasionally, zombies will throw axes, which destroy plants, so the player must regrow plants to maintain protection.
    \item \textbf{RealPortals:} is GVG-AI 2D adaptation of \emph{Portal} (Valve 2007). The player must reach the goal, which sometimes is behind a locked door that needs a key. Movement is restricted by water, which kills the player if they touch it. To succeed, players need to be creative in overcoming this hazard by using portals which can teleport them across the map. Players need to pick up wands, which allow them to toggle between the ability to create \emph{portal entrances} and \emph{portal exits}. There are also boulders on some levels, which the player can push into the water to transform the water into solid ground, creating land-bridges on which they can walk.
\end{itemize}

\section{Methods}
This project is a continuation of Khalifa et al.~\cite{khalifa2019intentional} where we are using Constrained MAP-Elites (CME) to illuminate the behavior characteristic search space. We are trying to find levels that are playable and at the same time are simple enough to work as game tutorials (it is clear what need to be done to finish that level). Similary to Khalifa et al.~\cite{khalifa2019intentional}, we are using the game mechanics as behavior characteristics for the CME algorithm. This process will end finding all the possible playable levels that have different combination of the game actual mechanics. In this work, we are calling this process ``Mechanic Illumination'' as we are using an CME (illumination algorithm) to illuminates the mechanic search space (behavior charactersitics).

CME starts by intially generating random levels that are used to populate the intial map. Levels are generated randomly using GVG-AI's random level generator class provided with the framework~\cite{khalifa2016general}. These levels are evaluated with respect to their constraints, fitness, and behavior characteristic. Based on the behavior characteristics, the correct cell in the map is selected and then the new level is inserted within. Levels are placed based on their constraints. If they satisfy the constraints, they are placed into feasible population, otherwise they are placed in the infeasible population. If the population is full, the new level will replace the worst agent within that population.

The consecutive population is generated by selecting a random cell from the MAP then we select a chromosome based on a predefined probability value from either the ``feasible'' or ``infeasible'' population. The new levels are generated by either applying our genetic operator on the selected level or generating it from GVG-AI's random level generator. This is to help the algorithm from potentially reaching a local optimum during generation. This process is repeated indefinitely to create the best levels for all the cells in the map.

\subsection{Level Representation}
The first difference between this work and the previous work~\cite{khalifa2019intentional} is the level representation. In Super Mario Bros, we used vertical slices that were sampled from the original Mario level. This representation is specific to Mario as levels are traversed from left to right allowing for fixed height levels. In this work, we represent the levels as 2D a array of tiles where each tile correspond to a certain game sprite. The differnt game tiles are extracted automatically from the VGDL description of the game. This representation is more generic and can work between different games with no needed modifications.

\subsection{Genetic operators}
In previous work, the mutation of the next population's levels were made using the crossover technique. However, we decided not to utilize crossover as it was harder to define a meaningful crossover in the new representation compared to the Mario vertical slice representaiton. The GVG-AI levels are represented as 2D ascii maps - generated with randomized dimensions. Selecting a portion of a large generated level for crossover could entirely erase the contents of another smaller level. Determining the crossover point and amount would also lead to level mutation inconsistencies. Our mutation operator selects a random tile of the level and turned into a random new tile value. Then, based on some set probability, another tile from the level is randomly chosen and mutated. This process continues until the probability check fails. The end result is a mutated version of the input level, ideally, yield a better fitted level. This mutated level is evaluated in the next iteration's population of levels before being added back to the MAP.

\subsection{Level Evaluation}
Levels are evaluated based on two parts: constraints and fitness. Similar to Khalifa et al.~\cite{khalifa2019intentional}, the constraint value focuses on providing an accessible playable level, while the fitness focuses on simplifying the levels so they can be easily parsed by humans. 

\subsubsection{Constraints}
The constraints of the level generation consisted of 2 parts: playability constraints and accessibility constraints. Playability constraints tries to make sure the level is playable and can be won in an appropriate amount of time called ``ideal time''. Accessibility constraints tries to make sure that the agent isn't defeated in the first few frames of the game. This allows the generated levels to be playable by humans, as players have enough time to react and won't immediately be defeated on level initialization. In this work, we use the AdrianCTX algorithm~\cite{perez2016general}, winner of the 2016 planning competition, as the evaluation agent. 

For the playability constraints, if the agent successfully completed the level, then only the completion time is evaluated. A preset value called the "ideal time" was used to compare to the completion time. The closer the completion time is to the ideal time, the better the constraint value. This is to ensure that the agent doesn't complete the level too fast - so the player cannot see the demonstration of the game's mechanics - or too slow - so the tutorial level doesn't drag out longer than needed. Otherwise if the agent does not win the level, the constraint value is inversely propotional to difference between the survival time and the ideal time. We multiply this value by $0.25$ to penalize it for not winning the level. Equation \ref{eq:constraint1} shows the part of the constraint calculation that applies to the time to complete a level,

\begin{equation}
P = \frac{win}{| T_{win} - T_{ideal} | + 1} \\
+ \frac{(1 - win) * 0.25}{|T_{survival} - T_{ideal} | + 1}
\label{eq:constraint1}
\end{equation}

where $win$ represents a 1 if the agent finished the level successfully and a 0 otherwise. $T_{ideal}$ represents the ideal time pre-defined for the system, and $T_{win}$ and $T_{survival}$ represent the finishing time of the agent before successful completion and unsuccessful completion respectively.

For the accessibility constraint, a "Do Nothing" agent is run on the level for a certain number of trials. This agent does not perform any user actions and remains idle in the level. If this agent dies before reaching $T_{ideal}$, the level fails the evaluation. If the agent does not survive for a majority of the evaluations tested, the ratio of successful idle agents over the number of times attempted is applied to the constraints. This is to remove the chance that the evaluation agent happened to "get lucky" on its performance in the level and keep the level reasonably user-friendly. Equation \ref{eq:constraint2} demonstrates how the idle agent's trials were applied,

\begin{equation}
    A = \begin{cases}
1 &\text{if $\frac{N_{pass}}{N_{total}} \ge 0.5$}\\
\frac{N_{pass}}{N_{total}} &\text{otherwise}
\end{cases}\\
\label{eq:constraint2}
\end{equation}

where $N_{pass}$ representing the number of times the idle agent survived to the ideal time, and $N_{total}$ representing the number of idle agent tests. 

The total constraint score of a level is therefore equivalent to \emph{C} in Equation \ref{eq:constraint_total},

\begin{equation}\label{eq:constraint_total}
    C = P + A
\end{equation}

where \emph{P} is from Equation \ref{eq:constraint1} and \emph{A} is from Equation \ref{eq:constraint2}. In order to be considered a ``feasible'' level for a MAP-Elites cell, the level must reach a certain threshold of constraints. If the level evaluation does not reach this threshold, the level is placed in the MAP-Elites cell's ``infeasible'' population instead.

The constraint threshold is chosen with the intention that a level will both be winnable and pass the idle agent tests, but may be within a certain range for $T_{ideal}$. For example, if the constraint threshold is set to 0.1, based on the function defined for the constraint value, a level can be considered elite in two cases: if it is winnable and passes the DoNothing tests, so long as $T_{win}$ is within 10 timesteps of $T_{ideal}$; or if it is unwinnable, but passes the DoNothing tests and $T_{survival}$ is within 2 timesteps of $T_{ideal}$. For our experiment, we set this threshold value to 0.1.

It is possible for an unwinnable level to be evaluated as a ``feasible'' level or winnable level to evaluated as an ``infeasible'' level. In the case of winninable levels, this could happen if the level has a poor finishing time (i.e. $T_{win}$ was not close enough to $T_{ideal}$) or the level is failing the idle agent tests. In the case of unwinnable levels, a level can still have a good constraint value (i.e. pass the constraint threshold) if $T_{survival}$ is extremely close to $T_{ideal}$ and also passes the idle tests. However, this end value will also be penalized to $0.25$ of the original value since the level was still not finished sucessfully. 


\subsubsection{Fitness}
The fitness of a level was determined by the tile entropy and the derivative-tile entropy of the level. We decided to use the same fitness function as the previous project~\cite{khalifa2019intentional} because we believe that simplisitic levels tend to be more aesthetically pleasing and enjoyable than ones that are noisy and chaotic. Minimalistic levels can also more clearly showcase meaningful elements within the level. All of this coincides with the motivation of this paper, which is to use this method to generate simple tutorial levels ~\cite{green2017press}. Minimizing entropy in a level will create fewer distractions for the player while they are playing the level and exploring different mechanics. The system evolves to create open, mostly empty-tiled levels or levels with similar tiles placed adjacent to each other that still demonstrate the game mechanics needed to play the game. Weights were given to the importance of the tile entropy versus the tile derivative entropy to create less noisy levels. Equation \ref{eq:fitness} was used for the level fitnesses: 
\begin{equation}
    fitness = H(lvl) * w + H(\Delta lvl) * (1 - w)
    \label{eq:fitness}
\end{equation}
where $H(lvl)$ represents the raw tile entropy of a level, $H(\Delta lvl)$ represents the entropy of the derivative of the same level, and $w$ represents the pre-set weight value. This equation was based on the entropy tile fitness equation used by \cite{khalifa2019intentional}, however, the level derivative is calculated differently. Since the level was not seperated into vertical slices and mutated on the slices like the Mario levels, the derivative was calculated based on the vertical and horizontal changes instead of only horizontal changes. For each tile, we calculate the number of different neighboring tiles on the north, south, east, and west of it use the value as the dervative value for the map.

\subsection{Behavior Characteristics}\label{sec:CME}

\begin{figure}
    \centering
    \begin{subfigure}[t]{.48\linewidth}
        \includegraphics[width=0.95\linewidth]{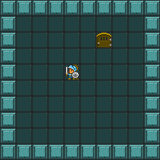}
        \caption{No 'get key' | No 'kill enemy'}
        \label{fig:idealzelda_a}
    \end{subfigure}
    \begin{subfigure}[t]{.48\linewidth}
        \includegraphics[width=0.95\linewidth]{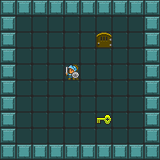}
        \caption{'get key' | No 'kill enemy'}
        \label{fig:idealzelda_b}
    \end{subfigure}
    \begin{subfigure}[t]{.48\linewidth}
        \includegraphics[width=0.95\linewidth]{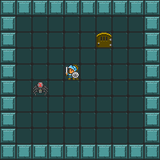}
        \caption{No 'get key' | 'kill enemy'}
        \label{fig:idealzelda_c}
    \end{subfigure}
    \begin{subfigure}[t]{.48\linewidth}
        \includegraphics[width=0.95\linewidth]{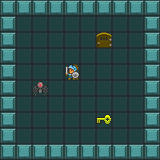}
        \caption{'get key' | 'kill enemy'}
        \label{fig:idealzelda_d}
    \end{subfigure}
    \caption{An example of an optimum MAP-Elites matrix for the Zelda game mechanics "get key" and "attack enemy".}
    \label{fig:dimensions}
\end{figure}

The behavior characteristic of each CME cell consists of multiple binary dimensions that correspond to the game mechanics. Each dimension represents whether or not a particular mechanic was performed by AdrianCTX agent that is used during calculating the constraints. For example, if the game mechanics for Zelda consisted of the list ``get key'' and ``kill enemy'', the CME behavioral characteristic will be 2 binary dimensions which will create 4 cells (~``get key'' \& ~``kill enemy'', No ``get key'' \& ``kill enemy'', ``get key'' \& ~``kill enemy'', and ``get key'' \& ``kill enemy''). Figure~\ref{fig:dimensions} shows the ideal levels for all these four possible cells. With this, the number of possible cells that can be generated for each game is $2^n$ with $n$ being the number of game mechanics defined for the game. The difference between this work and the previous work~\cite{khalifa2019intentional} is that these mechanics are automatically extracted from their VGDL description file instead of being provided by the humans.

\section{Experiments}
Four games from the GVG-AI framework were used to test the effectiveness of our level generation method. Zelda, Solarfox, Plants, and RealPortals are all described in Section \ref{sec:GVG-AI}. The dimensionalities used in the system are shown in Tables \ref{tab:zelda_mechanics}, \ref{tab:solarfox_mechanics}, \ref{tab:plants_mechanics}, and \ref{tab:realportals_mechanics}.
These mechanics were extracted from each game automatically by using the AtDelfi system~\cite{green2018atdelfi}, which is able to parse game rules directly from a GVG-AI game's VDGL description file. 

For all experiments, we generate 50 chromosomes for each iteration. In each iteration, 20\% of the levels are randomly initialized, and the remainder are filled with mutated versions of the selected chromosomes. For the Solarfox experiment, levels are much more dependent on having fewer empty tiles for functional gameplay and thus a less constrained initialized population. To assist evolution, only 10\% of Solarfox levels were randomly initialized each iteration, and 90\% filled with mutated versions. 

Each experiment ran for a total of 500 iterations. The dimensions for the level are calculated using the AdrianCTX agent's playthrough. The \emph{idle agent} is run a total of $5$ times on the level, of which it must survive $3$ in order to pass the constraint test. In RealPortals, it is impossible for a non-moving agent to die (the only way to lose is to fall into water), and therefore this constraint test is not necessary. Both agent's ``ideal times'' ($T_{ideal}$) were set to $70$ timesteps for all experiments.

If the constraint test is passed, the level's fitness is evaluated according to Equation \ref{eq:fitness}, where $w$ is 0.25 and $(1-w)$ is 0.75 in the Zelda and RealPortals experiments.  Plants and Solarfox levels are more dependent on tile uniformity and open areas, as opposed to Zelda and RealPortals. Therefore, $w$ was 0.2 while $(1-w)$ is 0.8 for thes experiments.  After evaluation, the chromosome is compared to its respective dimensional family, as specified in Section \ref{sec:CME}.

After evaluation, the system populates the feasible and infeasible populations within the MAP using the newly generated levels. For all four games, a single MAP-Elite cell is allowed to store a maximum of $20$ infeasible levels and $1$ ``feasible'' level which we call the ``elite'' level. A newly initialized level has a 50\% chance of being mutated from the elite level of a MAP-Elite cell. Otherwise, the level is mutated from the cell's best level from the infeasible population. 

\section{Results}

\begin{figure}
    \centering
    \includegraphics[width=.8\linewidth]{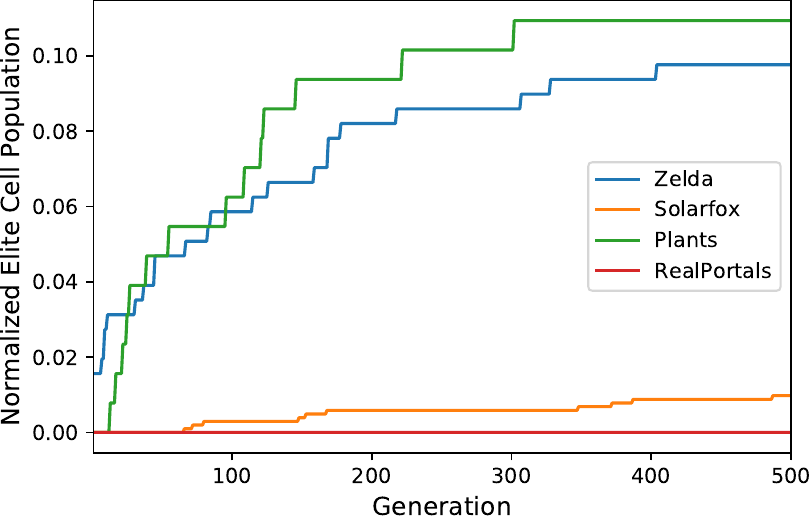}
    \caption{Number of filled Elite MAP Cells (Normalized) across generations. Although RealPortals is considerably lower than the other games, it's map contained 231 elites, nearly ten times more than any other game.}
    \label{fig:mapcell}
\end{figure}

The normalized elite counts across generations for our experiments is displayed in Figure \ref{fig:mapcell}. Each experiment was normalized against its total possible elite cell count, calculated using the game's mechanic dimensionality specified in Tables \ref{tab:zelda_mechanics}, \ref{tab:solarfox_mechanics}, \ref{tab:plants_mechanics}, and \ref{tab:realportals_mechanics}. 

\subsection{Mechanical Frequency in Elites}
\begin{figure*}
    \centering
    \begin{subfigure}[t]{.31\linewidth}
        \includegraphics[width=0.95\linewidth]{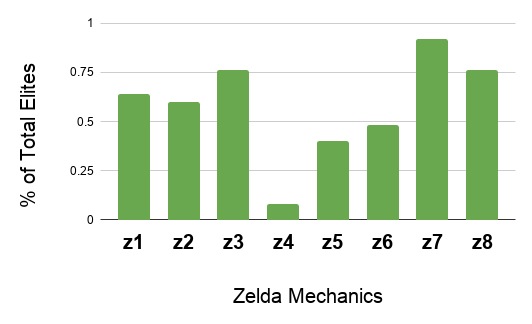}
        \caption{Zelda}
        \label{fig:zelda_mech_elites}
    \end{subfigure}
    \begin{subfigure}[t]{.31\linewidth}
        \includegraphics[width=0.95\linewidth]{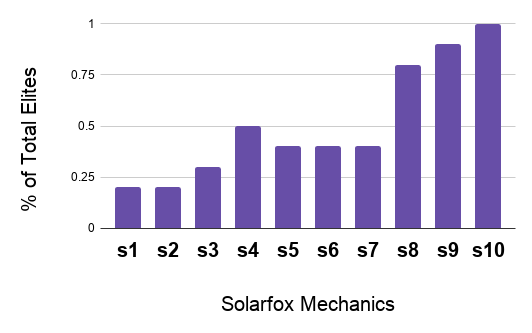}
        \caption{Solarfox}
        \label{fig:solarfox_mech_elites}
    \end{subfigure}
    \begin{subfigure}[t]{.31\linewidth}
        \includegraphics[width=0.95\linewidth]{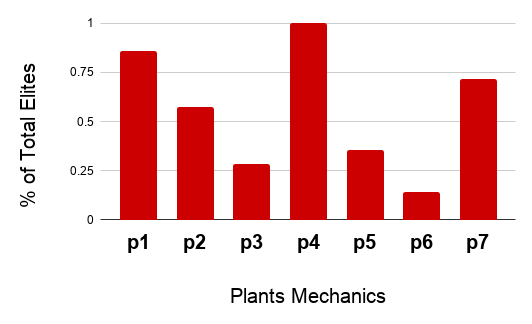}
        \caption{Plants}
        \label{fig:plants_mech_elites}
    \end{subfigure}
    \begin{subfigure}[t]{\linewidth}
        \includegraphics[width=0.95\linewidth]{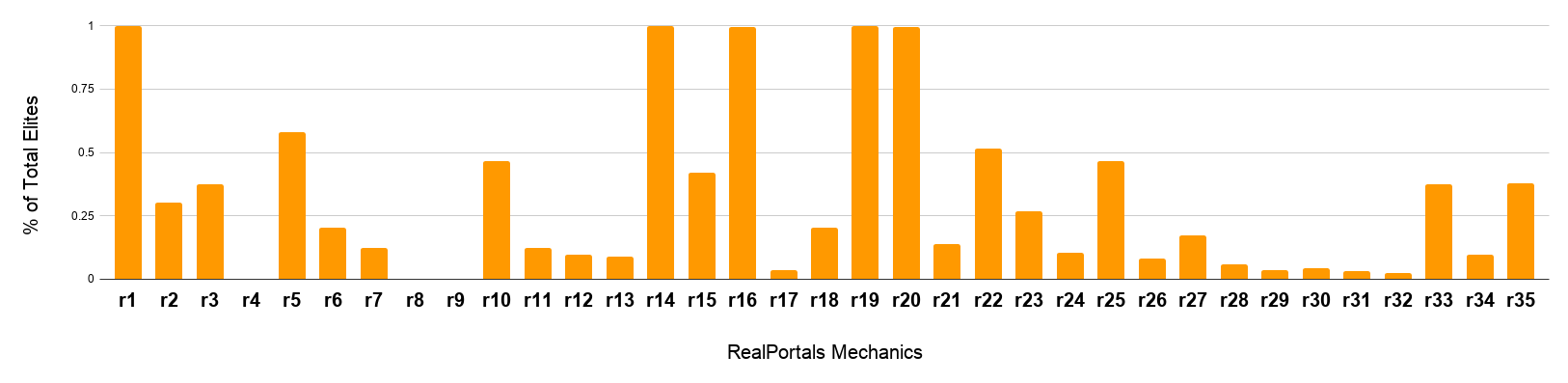}
        \caption{RealPortals}
        \label{fig:realportals_mech_elites}
    \end{subfigure}
    \caption{The percentage of elites that contain a specific mechanic for each game. The lettering of a mechanic corresponds to that games mechanic table. Zelda: Table \ref{tab:zelda_mechanics}; Solarfox: Table \ref{tab:solarfox_mechanics}; Plants: Table \ref{tab:plants_mechanics}; RealPortals: Table \ref{tab:realportals_mechanics}.}
    \label{fig:aggregate_mech_elites}
\end{figure*}

Figure \ref{fig:aggregate_mech_elites} displays the symbolicly represented mechanics present across all games and how prevalent each exists among that game's elite population. There are $3$ mechanics (d, h, and i) in RealPortals that are never expressed within any of the elites. These correspond to the ``drown,'' ``teleport-exit,'' and ``no-moving-boulder'' mechanics. The irony of the low activation of teleport mechanics (``teleport-entrance'' $= 8\%$, ``teleport-exit'' $= 0\%$) in a game called ``RealPortals'' is not lost on us, and this is further represented when looking at the elites themselves, which do not require teleportation to win. However, no agent has ever been submitted to the GVG-AI competition that can reliably beat RealPortals levels. The system's constraint function, which drives evolution to produce beatable levels, causes the generator to develop levels simple enough for the agent to win. Because teleportation drastically expands the space that the agent needs to search, the simplest solution for the generator is to remove the need to teleport.

\begin{table}[t]
    \centering
    \begin{tabular}{| p{0.04\linewidth} | p{.2\linewidth} | p{.6\linewidth} | }
    \hline
        * & Dimension & Description \\
    \hline \hline
        z1 & space-nokey & Agent pressed the space bar when the avatar did not have a key \\
    \hline
        z2 & space-withkey & Agent pressed the space bar when the avatar had a key \\
    \hline
        z3 & stepback & A sprite ran into another sprite \\
    \hline
        z4 & kill-nokey & A sprite killed the avatar when the avatar did not have a key \\
    \hline
        z5 & kill-withkey & A sprite killed the avatar when the avatar had a key \\
    \hline
        z6 & sword-kill & The agent killed an enemy sprite with a sword \\
    \hline
        z7 & getkey & The agent picked up a key \\
    \hline
        z8 & touchgoal & The agent touched a goal with the key and won the game \\ 
    \hline
    \end{tabular}
    \caption{Constrained MAP-Elites dimensions for the GVG-AI game Zelda}
    \label{tab:zelda_mechanics}
\end{table}

\begin{figure}[b]
    \centering
    \begin{subfigure}[t]{.48\linewidth}
        \includegraphics[width=0.95\linewidth]{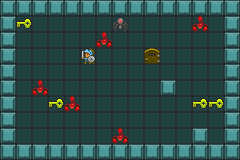}
        \caption{mean: z1, z2, z3, z5, z7}
        \label{fig:zelda_levels_a}
    \end{subfigure}
    \begin{subfigure}[t]{.48\linewidth}
        \includegraphics[width=0.95\linewidth]{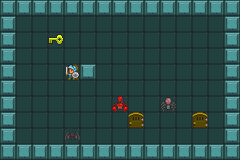}
        \caption{mode: z1, z2, z3, z7, z8}
        \label{fig:zelda_levels_b}
    \end{subfigure}
    \begin{subfigure}[t]{.48\linewidth}
        \includegraphics[width=0.95\linewidth]{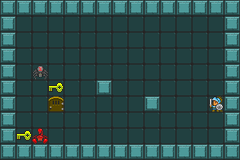}
        \caption{least: z7, z8}
        \label{fig:zelda_levels_c}
    \end{subfigure}
    \begin{subfigure}[t]{.48\linewidth}
        \includegraphics[width=0.95\linewidth]{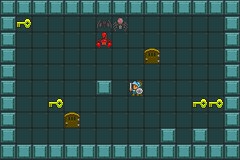}
        \caption{most: z1, z2, z3, z5, z6, z7, z8}
        \label{fig:zelda_levels_d}
    \end{subfigure}
    \caption{A subset of generated elite levels for Zelda. Their string representation corresponds to their showcased mechanics detailed in Table \ref{tab:zelda_mechanics}.}
    \label{fig:zelda_levels}
\end{figure}

\subsection{In-Depth Game Analysis}
In the following subsections, we present a representative subset of each game's generated levels. The \emph{mean} and \emph{mode} levels correspond to the mean and mode number of mechanics triggered across all elites. When multiple elites contained the identical amount of mechanics for either mean or mode, we randomly sampled among these elites to display one of them. We realize that this is only a subset of the possible elites, and that dimensional similarity does not necessarily equate to structural similarity.

\subsubsection{Zelda}
After 500 iterations, 55 out of 256 possible cells were populated for the Constrained MAP-Elites matrix of Zelda, with 25 cells containing an elite map. The average fitness for these cells was 0.5186. Figure \ref{fig:zelda_levels} displays four elites at opposite ends of the dimensional spectrum. The mean and mode elites are calculated to have $4.64$ and $5$ dimensions. respectively. The map with the least dimensionality (Figure \ref{fig:zelda_levels_c}) showcased $2$ mechanics, out of a possible 8. The map with the most mechanic-dimensionality~\ref{fig:zelda_levels_d} contained $7$.

The least dimensional map matches up with the AtDelfi system's critical mechanics~\cite{green2018atdelfi}. We can see however, that other mechanics from Table \ref{tab:zelda_mechanics} can be triggered in this space, such as killing monsters and bumping into walls. We think due to the wide-openess of the space, the agent failed to bump into any walls or monsters on its bee-line route to the key and the door. The most dimensional map showcases nearly all possible mechanics, only missing kill-nokey. We can interpret this to mean the agent went for the key first, before dealing with monsters. At a glance, it would be possible to also trigger the kill-nokey mechanic, depending on agent priority and the monster movement patterns. 

\begin{table}[t]
    \centering
    \begin{tabular}{| p{0.04\linewidth} | p{.2\linewidth} | p{.6\linewidth} | }
    \hline
        * & Dimension & Description \\
    \hline \hline
        s1 & hit-wall & Agent hit a wall \\
    \hline
        s2 & hit-enemyground & Agent touched enemy ground \\
    \hline
        s3 & hit-avatar & An enemy sprite hit the Agent \\
    \hline
        s4 & touch-powerblib & The agent touched a powerblib \\
    \hline
        s5 & spawn-more & A turning powerblib created a blib \\
    \hline
        s6 & change-blib & A turning powerblib changed into a normal blib \\
    \hline
        s7 & overlap-blib & A powerblib overlapped with a blib \\
    \hline
        s8 & get-blib & The agent got a blib\\ 
    \hline
        s9 & reverse-direction & A sprite hit a wall and reversed direction \\ 
    \hline
        s10 & enemy-shoot & An enemy fired a missile at the avatar \\ 
    \hline
    \end{tabular}
    \caption{Constrained MAP-Elites dimensions for the GVG-AI game Solarfox}
    \label{tab:solarfox_mechanics}
\end{table}

\begin{figure}[b]
    \centering
    \begin{subfigure}[t]{.48\linewidth}
        \includegraphics[width=0.95\linewidth]{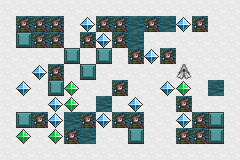}
        \caption{mean: *b*d*hij}
        \label{fig:solarfox_levels_a}
    \end{subfigure}
    \begin{subfigure}[t]{.48\linewidth}
        \includegraphics[width=0.95\linewidth]{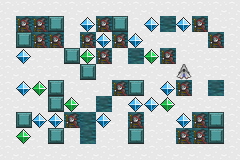}
        \caption{mode: *c*ij}
        \label{fig:solarfox_levels_b}
    \end{subfigure}
    \begin{subfigure}[t]{.48\linewidth}
        \includegraphics[width=0.95\linewidth]{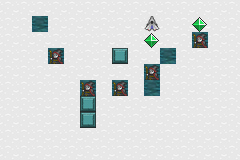}
        \caption{least: *hij}
        \label{fig:solarfox_levels_c}
    \end{subfigure}
    \begin{subfigure}[t]{.48\linewidth}
        \includegraphics[width=0.95\linewidth]{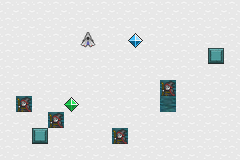}
        \caption{most: *cdefghij}
        \label{fig:solarfox_levels_d}
    \end{subfigure}
    \caption{A subset of generated elite levels for Solarfox. The string representation corresponds to their showcased mechanics detailed in Table \ref{tab:solarfox_mechanics}.}
    \label{fig:solarfox_levels}
\end{figure}

\subsubsection{Solarfox}
After 500 iterations, 52 out of 1024 possible cells were populated for the Constrained MAP-Elites matrix of Solarfox game, with 10 cells containing an elite map. The average fitness across all cells was 0.4311. Figure \ref{fig:solarfox_levels} displays four elites at opposite ends of the dimensional spectrum. The mean and mode are calculated to be $5.1$ and $3$ respectively. The least mechanic-dimensional map contained only $3$ mechanics (Figure \ref{fig:solarfox_levels_c}), whereas the most ~ (Figure \ref{fig:solarfox_levels_d}) contained $8$ out of the possible $10$ mechanics. 

The representative least dimensional elite (Figure \ref{fig:solarfox_levels_c} presents a lightly populated level containing just a few enemies, blibs, and walls. It contains no powerblib-generators, only normal blibs, which are placed incredibly close to the player at start for an easy win. The most dimensional elite contains nearly every mechanic in the game except for $2$: the agent hitting a wall and touching enemy ground tile where both kills the agent upon touching them.

\begin{table}[t]
    \centering
    \begin{tabular}{| p{0.04\linewidth} | p{.2\linewidth} | p{.6\linewidth} | }
    \hline
        * & Dimension & Description \\
    \hline \hline
        p1 & space & Agent pressed the SpaceBar \\
    \hline
        p2 & hit-wall & A sprite touched a wall \\
    \hline
        p3 & kill-plant & A zombie destroyed a plant \\
    \hline
        p4 & zombie-goal & A zombie sprite reaches the goal \\
    \hline
        p5 & pea-hit & A pea sprite hits a zombie sprite \\
    \hline
        p6 & tomb-block & A tomb sprite blocks a pea sprite \\
    \hline
        p7 & make-plant & Agent placed a plant on a marsh tile \\
    \hline
    \end{tabular}
    \caption{Constrained MAP-Elites dimensions for the GVG-AI game Plants}
    \label{tab:plants_mechanics}
\end{table}

\begin{figure}[h]
    \centering
    \begin{subfigure}[t]{.48\linewidth}
        \includegraphics[width=0.95\linewidth]{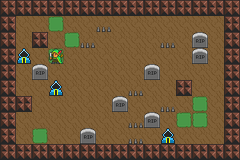}
        \caption{mean: p1, p3, p4, p5, p7}
        \label{fig:plants_levels_a}
    \end{subfigure}
    \begin{subfigure}[t]{.48\linewidth}
        \includegraphics[width=0.95\linewidth]{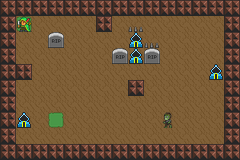}
        \caption{mode: p5, p6, p7}
        \label{fig:plants_levels_b}
    \end{subfigure}
    \begin{subfigure}[t]{.48\linewidth}
        \includegraphics[width=0.95\linewidth]{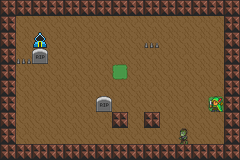}
        \caption{least: p4}
        \label{fig:plants_levels_c}
    \end{subfigure}
    \begin{subfigure}[t]{.48\linewidth}
        \includegraphics[width=0.95\linewidth]{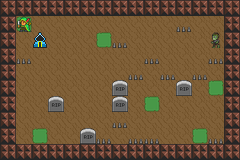}
        \caption{most: p1, p2, p3, p4, p5, p7}
        \label{fig:plants_levels_d}
    \end{subfigure}
    \caption{A subset of generated elite levels for Plants. Their string representation corresponds to their showcased mechanics detailed in Table \ref{tab:plants_mechanics}.}
    \label{fig:plants_levels}
\end{figure}

\subsubsection{Plants}

After 500 iterations, 31 out of 128 possible cells were filled for the Constrained MAP-Elites matrix of Plants, with 14 cells containing an elite map. The average fitness across all cells was 0.3993. Figure \ref{fig:plants_levels} displays four elites of varying dimensions. The mean and mode are calculated to be $3.93$ and $5$ respectively. The map with the least dimensionality showcased just $1$ mechanic. The map with the most mechanic-dimensionality contained $6$ out of the $7$ possible mechanics.

Unlike any of the other elites of any other game, the representative least dimensional elite of Plants contains a single mechanic, which happens to be one that causes the player to lose the game. Based on the game rules, it is not possible to win this level no matter what actions the player does, as the zombies will spawn several tiles to the right of the villager and inevitably collide with it. We think that the algorithm optimized the zombie spawner placement such that it takes almost $T_{survival}$ before losing which will also allow the idle agent to pass almost all its trials. The elite with the most activated dimensions, on the other hand, was possible to win. The triggered mechanics guarantee that a player could encounter most of the mechanics in the game during play. 

\begin{figure}
    \centering
    \begin{subfigure}[t]{.48\linewidth}
        \includegraphics[width=0.95\linewidth]{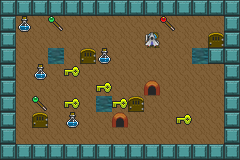}
        \caption{mean: r1, r3, r5, r11, r14, 15, r16, r19, r20, r21, r24}
        \label{fig:realportals_levels_a}
    \end{subfigure}
    \begin{subfigure}[t]{.48\linewidth}
        \includegraphics[width=0.95\linewidth]{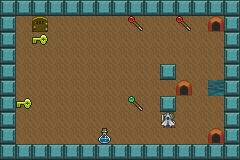}
        \caption{mode: r1, r3, r5, r13, r15, r17, r18, r19, r20, r21, r22}
        \label{fig:realportals_levels_b}
    \end{subfigure}
    \begin{subfigure}[t]{.48\linewidth}
        \includegraphics[width=0.95\linewidth]{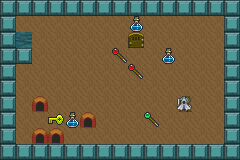}
        \caption{least: r1, r3, r14, r19, r22}
        \label{fig:realportals_levels_c}
    \end{subfigure}
    \begin{subfigure}[t]{.48\linewidth}
        \includegraphics[width=0.95\linewidth]{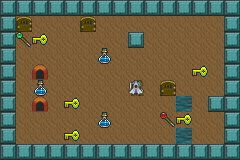}
        \caption{most: r1, r5, r7, r10, r11, r13, r14, r15, r16, r19, r20, r23, r25, r26, r34, r35}
        \label{fig:realportals_levels_d}
    \end{subfigure}
    \caption{A subset of generated elite levels for realportals. Their string representation corresponds to their showcased mechanics detailed in Table \ref{tab:realportals_mechanics}.}
    \label{fig:realportals_levels}
\end{figure}
\begin{table}[]
     \centering
     \resizebox{\linewidth}{!}{
     \begin{tabular}{| p{0.04\linewidth} | p{.35\linewidth} | p{.65\linewidth} | }
     \hline
         * & Dimension & Description \\
     \hline \hline
        r1 & space & Agent pressed the SpaceBar \\
     \hline
        r2 & change-key-blue & changes blue avatar's current resource to a key \\
     \hline
        r3 & hit-wall & A sprite touched a wall \\
     \hline
        r4 & drown & destroy any sprite that falls in the water \\
     \hline
        r5 & toggle-blue & avatar changes current portal shot to blue \\
     \hline
        r6 & no-lock & any sprite tried to move through a lock \\
     \hline
        r7 & no-portalexit & any sprite tried to move through an exit portal \\
     \hline
        r8 & teleport-exit & orange avatar steps through the entrance portal \\
     \hline
        r9 & no-moving-boulder & sprite tried to move through a moving boulder \\
     \hline
        r10 & no-idle-boulder & sprite tried to move through an idle boulder \\
     \hline
        r11 & change-key-orange & changes orange avatar's current resource to a key \\
     \hline
        r12 & toggle-orange & avatar changes current portal shot to orange \\
     \hline
        r13 & teleport-entrance & blue avatar steps through exit portal \\
     \hline
        r14 & get-weapon & avatar picks up a weapon \\
     \hline 
        r15 & get-key & avatar picks up a key \\
     \hline
        r16 & back-to-wall & portal turns back into a wall \\
     \hline
        r17 & fill-water & moving boulder falls into water to fill it \\
     \hline
        r18 & open-lock & avatar unlocks a lock \\
     \hline
        r19 & touchgoal & The agent touched a goal and won the game \\ 
     \hline
        r20 & make-portal & wall turns into a portal \\
     \hline 
        r21 & portal-missile-velocity & send a missile through a portal at the same velocity \\
     \hline
        r22 & cover-goal & goal is covered by a missile \\
     \hline
        r23 & blue-missile-in & send a blue missile thru a portal entrance \\
     \hline 
        r24 & orange-missile-in & send an orange missile thru a portal entrance \\
     \hline
        r25 & portal-boulder & send a boulder through a portal at the same velocity \\
     \hline
        r26 & stop-boulder-key & moving boulder stops after hitting a key \\
     \hline
        r27 & stop-boulder-wall & moving boulder stops after hitting a wall \\
     \hline
        r28 & stop-boulder-blue-toggle & moving boulder stops after hitting a blue portal toggle \\
     \hline
        r29 & stop-boulder-orange-toggle & moving boulder stops after hitting an orange portal toggle \\
     \hline
        r30 & stop-boulder-lock & moving boulder stops after hitting a lock \\
     \hline
        r31 & teleport-boulder & sends a boulder to the other portal \\
     \hline
        r32 & stop-boulder-boulder & moving boulder stops after hitting another boulder \\
     \hline
        r33 & stop-boulder-avatar-blue & moving boulder stops after hitting the blue avatar \\
     \hline
        r34 & stop-boulder-avatar-orange & moving boulder stops after hitting the orange avatar \\
     \hline
        r35 & roll-boulder & boulder moved over a tile \\
     \hline
     \end{tabular}
     }
     \caption{Constrained MAP-Elites dimensions for the GVG-AI game RealPortals}
     \label{tab:realportals_mechanics}
 \end{table}

\subsubsection{RealPortals}

After 500 iterations, 6966 cells were filled for the Constrained MAP-Elites matrix of RealPortals, with 231 cells containing an elite map. The average fitness across all cells was 0.4257. Figure \ref{fig:realportals_levels} displays four elites of varying dimensions. The mean and mode are calculated to be $10.78$ and $10$ respectively. The least mechanic-dimensional map contained $5$ mechanics, and the most contained $16$ out of a possible $35$.

In contrast to the other games described above, RealPortals is extremely complex, echoed in the sheer amount of elites. Ironically, the generated levels are all extremely simple to solve, unlike any of the GVG-AI included levels. Without water dividing the map and requiring the player to use portaling, the levels are transformed into a simple find-the-goal game even simpler than Zelda. Even if the agent uses a portal, there is no need to do so or to use any of the other game mechancis which are normally required by the framework game levels (pushing boulders into water, unlocking the lock with a key, etc). This is due to the inability of AdrianCTX agent of solving any level more complex than the one it puts into the cell (AdrianCTX can't solve normal RealPortal levels). The least dimensional representative elite still activates five mechanics (with others still possible, just not activated during playthrough), whereas the most dimensional elite can be beaten by taking two steps to the right.

\section{Discussion}\label{sec:discussion}
Constrained MAP-Elites was able to populate more than 10\% and slighly less than 10\% of the total cells with elites for Plants and Zelda respectively. Compared to Solarfox (approx. 1\%) and RealPortals(>1\%), these two games' dimensions were relatively well-explored. At first glance, it would make sense that RealPortals was not as explored, due to its $34$-dimensional complexity compared to Zelda's meager $8$-dimensions. However, Solarfox ($10$) is also many less dimensional than RealPortals, but has a similarly low relative elite population. Keep in mind that though the dimensionality of Solarfox relative to Zelda is only 2 higher, the dimensional space increases from 256 possible cells to 1024. We also hypothesize that elite population is impacted not only by the total number of game mechanics, but also by the ability of the agent to solve the level as AdrianCTX is unable to beat complex Solarfox levels~\cite{bontrager2016matching}. The reason of the bad perfomance on most levels due to game mechanics that allows for continous movement of the player's avatar. This mechanic forces the agent to be responsive at each frame as the agent can die quickly by not taking actions compared to the other games.

Across all games, when compared to the original levels, the generated levels provide a sense of uniformity. Solarfox levels tend to be sparsely populated with blibs, with a few exceptions (Figure \ref{fig:solarfox_levels_b}) instead of gemotric arrangements of blibs and powerblibs. There tend to be no water tiles present in RealPortals, or marsh tiles in Plants, relative to each games' original levels. The Zelda elites consist of wide open spaces, instead of the usual maze-like patterns. We hypothesize that all of this is due to the entropy pressure from the fitness calculation specified in Equation \ref{eq:fitness}, which drives evolution towards creating simplistic levels with large amount of empty tiles or highly populated levels filled with a lot of similar tiles.

Because of how the fitness is defined and dimensions are calculated, any activated mechanic on a map is guaranteed to have the \emph{possibility} of occuring during a playthrough. However, this does not guarantee that the mechanic \emph{must} be activated or that \emph{other mechanics} do not have the possibility of happening at all. For example: the least mechanical level of Zelda (figure~\ref{fig:zelda_levels_c}) only trigers two mechanics (``getkey'' and ``touchgoal'') although it has enemies and walls which can trigger mechanics their corresponding mechanics (``stepback'', ``kill-nokey'', and ``kill-withkey''). To guarantee either of these, the system would have to exhaustively search all possible game states of the level, which is not computationally feasible within a timely manner for any of these games.

We noticed that better generated levels depends on having better playing agents. Most of the GVGAI general players tends to not perform very well. It would be interesting to try to have an agent that adapt to each game using some game specific information. One way to do this might be to take advantage of hypterstate information~\cite{cook2019hyperstate}. Another way would be use the information extracted from the VGDL file as an evaluation function for the agent~\cite{green2019automatic}. Another direction would be to aggregate the unique mechanics triggered across a multitude of agents, to get a better sense of the mechanic space.

\section{Conclusion}
Outside of actual gameplay, isolating mechanics from each other to allow players to examine the full breadth of each game mechanic is a non-trivial problem. Our system looks to solve this issue by generating levels that are constructed using a bottom-up approach, evolving levels that are beatable and simple, while using the illuminating power of MAPElites to categorize levels by mechanics triggered. Using our method, one can generate a sampling of the possible mechanic combinations for four GVG-AI games and examine the behavior of their evolution throughout each generation as well as which game mechanics were critical to gameplay. 

This system serves as a proof of concept for developing isolated mechanic levels and could be beneficial in developing tutorial levels where mechanics would need to be demonstrated individually or in combination in a controlled environment. This would go hand-in-hand with an automatic tutorial generation system, which could use these levels for player practice. This system could also be beneficial in examining the minimal level structure needed for a game mechanic. The level design for each mechanic representation could be based their architechture on the MAPElites cells generated from the pre-set list of game mechanics.

In future work, we would like to test our system on more games from the GVG-AI framework, such as games with fewer mechanic combinations needed but with more variability in the level structure as well as games with more complex series of mechanic combinations needed to win the game (i.e Frogs or Sokoban.) Expansion of the system to games outside of the framework would also help to prove the usability and generability of the system for games that do not have a predefined description language.

As the core motivation of this paper is to expand the generation of experiences for tutorials, we used a minimizing entropy pressure to produce simplistic levels. One could imagine simplistic levels as being the ``beginner levels,'' the levels that a player first plays to learn about a mechanic. However, to achieve true mastery of a skill, a player must practice it and be challenged in a variety of difficult situations. Therefore, we would like to propose future work in which we experiment with a ``targeted entropy'' instead of a minimizing one. This would allow us to experiment with different level simplicities to create more difficult levels to beat.

\begin{acks}
Ahmed Khalifa acknowledges the financial support from NSF grant (Award number 1717324 - ``RI: Small: General Intelligence through Algorithm Invention and Selection.''). Michael Cerny Green and Megan Charity acknowledge the financial support of the SOE Fellowship from NYU Tandon School of Engineering.
\end{acks}

\bibliographystyle{ACM-Reference-Format}
\bibliography{sample-base}

\end{document}